# Accepted version

**Graphical abstract:**

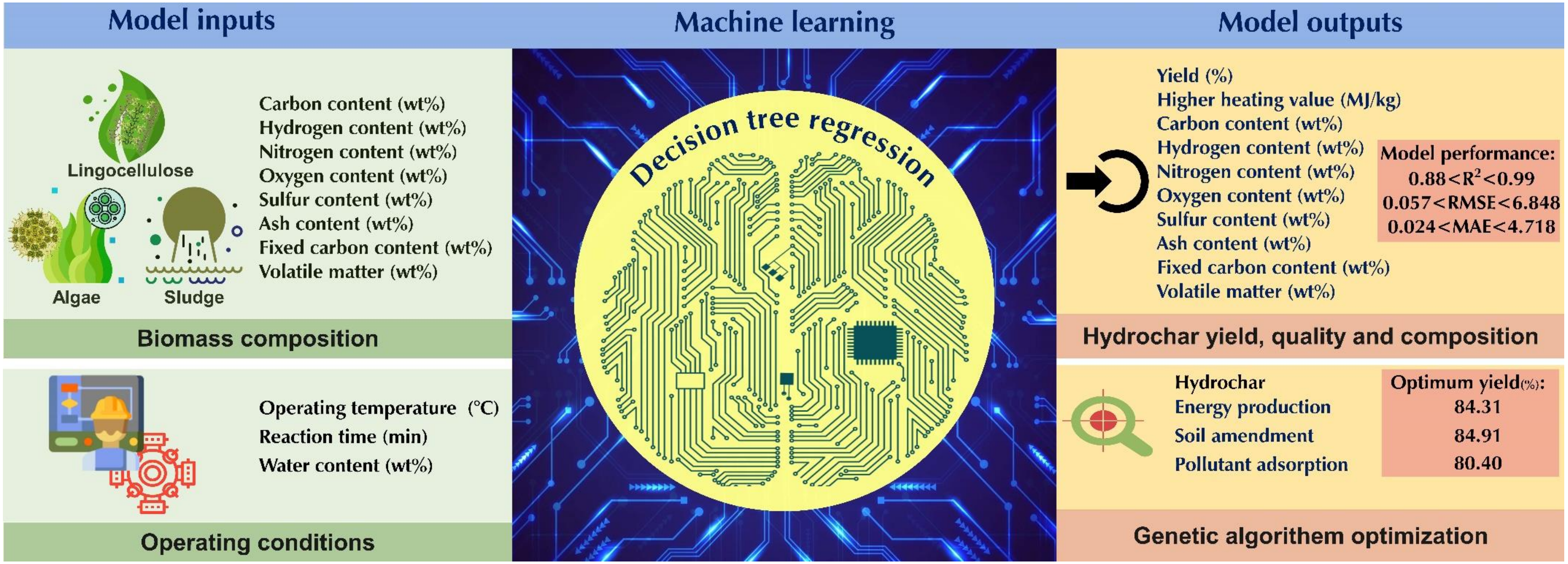

# Machine learning-based characterization of hydrochar from biomass: implications for sustainable energy and material production

Alireza Shafizadeh[1,2,†], Hossein Shahbeik[1,†], Shahin Rafiee[2], Aysooda Moradi[2], Mohammadreza Shahbaz[2], Meysam Madadi[3], Cheng Li[1], Wanxi Peng[1,*], Meisam Tabatabaei[4,5,1,*], Mortaza Aghbashlo[2,1,*]

[1] Henan Province Engineering Research Center for Forest Biomass Value-added Products, School of Forestry, Henan Agricultural University, Zhengzhou, 450002, China

[2] Department of Mechanical Engineering of Agricultural Machinery, Faculty of Agricultural Engineering and Technology, College of Agriculture and Natural Resources, University of Tehran, Karaj, Iran

[3] Key Laboratory of Industrial Biotechnology, Ministry of Education, School of Biotechnology, Jiangnan University, Wuxi 214122, China

[4] Higher Institution Centre of Excellence (HICoE), Institute of Tropical Aquaculture and Fisheries (AKUATROP), Universiti Malaysia Terengganu, 21030 Kuala Nerus, Terengganu, Malaysia

[5] Department of Biomaterials, Saveetha Dental College, Saveetha Institute of Medical and Technical Sciences, Chennai 600 077, India

†This author contributed equally.

*Corresponding authors:
Mortaza Aghbashlo (maghbashlo@ut.ac.ir), Meisam Tabatabaei (meisam_tab@yahoo.com)
Wanxi Peng (pengwanxi@henau.edu.cn)

**Abstract**

Hydrothermal carbonization (HTC) is a process that converts biomass into versatile hydrochar without the need for prior drying. The physicochemical properties of hydrochar are influenced by biomass properties and processing parameters, making it challenging to optimize for specific applications through trial-and-error experiments. To save time and money, machine learning can be used to develop a model that characterizes hydrochar produced from different biomass sources under varying reaction processing parameters. Thus, this study aims to develop an inclusive model to characterize hydrochar using a database covering a range of biomass types and reaction processing parameters. The quality and quantity of hydrochar are predicted using two models (decision tree regression and support vector regression). The decision tree regression model outperforms the support vector regression model in terms of forecast accuracy ($R^2 > 0.88$, RMSE < 6.848, and MAE < 4.718). Using an evolutionary algorithm, optimum inputs are identified based on cost functions provided by the selected model to optimize hydrochar for energy production, soil amendment, and pollutant adsorption, resulting in hydrochar yields of 84.31%, 84.91%, and 80.40%, respectively. The feature importance analysis reveals that biomass ash/carbon content and operating temperature are the primary factors affecting hydrochar production in the HTC process.

Abbreviations & Symbols

| | |
|---|---|
| $d_i$ | Difference between the two ranks of each observation |
| DTR | Decision tree regression |
| HHV | Higher heating value |
| HTC | Hydrothermal carbonization |
| SHAP | Shapely additive explanation |
| MAE | Mean absolute error |
| ML | Machine learning |
| $n$ | Number of data points |
| $n$ | Quantity of observation |
| $r$ | Spearman's rank correlation coefficient |
| $R^2$ | Coefficient of determination |
| RMSE | Root-mean-square error |
| SVR | Support vector regression |
| $\bar{y}$ | Average of the data |
| $y_a$ | True data |
| $y_p$ | ML-predicted data |

## 1. Introduction

Most of the world's energy and chemical needs are met by fossil fuels today. Researchers are exploring alternative energy and chemical sources due to the depletion of fossil reserves and their detrimental impact on the environment, including air pollution and climate change [1]. Among various renewable energy resources, carbon-neutral biomass energy (bioenergy) has received more attention over the last decades [2] because of its capability to be upgraded into various fossil-like solid, liquid, and gaseous fuels with a wide variety of applications. The most important renewable energy source on the planet is bioenergy, which makes up approximately 10% of the global primary energy supply [3]. About half of all biomass consumption worldwide goes toward cooking and heating in developing countries [4]. To this end, switching from traditional biomass combustion to more efficient modern energy forms such as biogas, biodiesel, bioethanol, and biopower with minimal negative effects on the local environment is essential.

Various thermochemical and biological technologies have been introduced to convert biomass into biofuels and biochemicals [5]. The scientific community has revealed a great interest in thermochemical conversion technologies like combustion, gasification, pyrolysis, hydrothermal liquefaction, and hydrothermal carbonization (HTC) [6,7]. Indeed, biomass feedstocks can be directly converted into value-added products without undergoing energy-intensive or chemically demanding pretreatment processes. However, owing to their high moisture content, most dedicated and waste biomass feedstocks cannot be directly handled by some thermochemical pathways, *i.e.*, combustion, gasification, and pyrolysis [8]. The inadequacy of these thermochemical methods for directly processing wet biomass feedstocks incurs a high cost to reduce their moisture content to the desired level. Hydrothermal processing technologies, like HTC, are an efficient way to avoid the energy-intensive drying step. This process also consumes less energy than pyrolysis and gasification [9]. Because of its milder reaction conditions, the HTC process produces few toxic gases, such as nitrogen and sulfur oxides.

The HTC process is carried out in hot pressurized water at temperatures ranging from 180 to 260 °C. This process resembles the natural coalification process that occurred millions of years ago. In minutes to hours, the HTC process produces a solid carbon-rich product, *i.e.*, hydrochar [10]. The resultant hydrochar has a fuel quality similar to lignite. Besides HTC, char can be produced from biomass pyrolysis (biochar). Biochar has high carbon content, high surface area, and low reactivity, making it a promising candidate for carbon sequestration, water treatment, and agriculture applications [11]. However, hydrochar has low carbon content, high ash content, and high reactivity [12,13], making it a promising material for energy generation, soil amendment, and pollutant adsorption [11]. Hydrochar is produced under relatively mild conditions compared to biochar, resulting in lower stability, lower porosity, smaller specific surface area, and lower pH [14]. Hydrochar is rich in oxygenated functional groups, such as carboxyl, carbonyl, hydroxyl, and phenolic hydroxyl, which facilitate metal adsorption [15].

The physicochemical and structural properties of hydrochar rely heavily on biomass properties and reaction processing parameters. However, identifying the optimal processing parameters for a particular biomass feedstock requires numerous trial-and-error experiments. Such experiments are time-consuming, labor-intensive, and costly, making it challenging to precisely characterize the quality and quantity of hydrochar derived from biomass HTC. Furthermore, existing computational approaches cannot adequately predict the properties of hydrochar due to the complex governing mechanisms of the HTC process. Therefore, there is a critical need for more efficient and effective methods to determine the physicochemical and structural characteristics of hydrochar, ensuring its suitability for various applications.

It is possible to address the abovementioned issues by employing advanced data-science tools like machine learning (ML) technology, which can unravel, quantify, and understand large and complicated datasets in bioenergy production systems [16]. By encapsulating statistical and mathematical algorithms, ML techniques can generalize important trends from existing historical data and predict values in unexplored space based on the trends of the training data [17]. The data-driven methods, however, require a tremendous amount of data to correlate input variables with desired outcomes. With the accumulated experimental data in the published literature, ML techniques can now effectively simulate the HTC process of biomass feedstocks without knowing the governing laws. Only a few studies have explored the quality and quantity parameters of hydrochar produced from the biomass HTC process using ML technology [18–20]. The characteristics and limitations of ML models developed to characterize hydrochar are presented in Table 1. According to these studies, ML could successfully characterize hydrochar with no prior knowledge of the mechanisms of biomass HTC.

**Table 1**. ML models developed for characterizing hydrochar.

| Feedstock(s) | Model type(s) | Data number | Independent variables | Dependent response(s) | Drawbacks | Ref. |
|---|---|---|---|---|---|---|
| Sewage sludge | Feed-forward neural network | 138 | CHNO analysis, proximate analysis (fixed carbon, ash content, volatile matter), reaction temperature, reaction time, water/biomass ratio | Nitrogen content in hydrochar | -Applicable only to sewage sludge -Applicable only to predicting the nitrogen content of hydrochar | [19] |
| Sewage sludge, food waste, manure | Support vector regression, random forest | 248 | CHNO analysis, proximate analysis (fixed carbon, ash content, volatile matter), reaction temperature, reaction time, water/biomass ratio | Hydrochar yield, hydrochar higher heating value, energy recovery, energy densification | -Applicable only to limited waste biomass feedstocks -Ability to predict a few hydrochar quality parameters | [18] |
| Sewage sludge, food waste, manure | Support vector regression, random forest | 248 | CHNO analysis, proximate analysis (fixed carbon, ash content, volatile matter), reaction temperature, reaction time, water/biomass ratio | Hydrochar yield, hydrochar higher heating value, energy recovery, hydrochar carbon content, carbon recovery, hydrochar N/C, H/C, and O/C ratio | -Applicable only to limited waste biomass feedstocks | [20] |
| Lignocellulosic biomass | Extreme gradient boosting method, multilayer perceptron artificial neural network, support vector machine | 81 | Cellulose, hemicellulose, lignin, water/biomass ratio, temperature, time | Hydrochar yield | -Applicable only to lignocellulosic biomass -Developed based on limited data patterns -Applicable only to predicting hydrochar yield | [21] |
| Sewage sludge and lignocellulosic | Extreme gradient boosting, random forest | 221 | CHSNO analysis and proximate analysis (fixed carbon, ash content, volatile matter) of both sewage | Carbon content, H/C ratio, O/C, N/C, HHV, fuel ratio, mass yield, | -Applicable only to co-HTC of sewage sludge and | [22] |
| biomass (co-HTC) | | | sludge and lignocellulosic biomass, reaction temperature, reaction time, solid loading, sewage sludge/lignocellulosic biomass ratio | energy yield of hydrochar | lignocellulosic biomass | |
| Municipal sludge | Random forest, gradient boosting tree, and artificial neural network | 246 | Total solid, organic matter, ash content, carbon content, H/C ratio, O/C ratio, N/C ratio, and HHV of municipal sludge, water content, temperature, reaction time | Carbon recovery, energy recovery, hydrochar HHV | -Applicable only to municipal sludge<br>-Ability to predict a few hydrochar quality parameters | [23] |
| Wood biomass, herbaceous biomass, food waste | Artificial neural network combined with particle swarm optimization | 296 | CHSNO analysis, proximate analysis (fixed carbon, ash content, volatile matter), temperature, holding time, solid-liquid ratio | Mass yield, hydrochar HHV, hydrochar ash content, hydrochar N/C ratio, dehydration degree, decarboxylation degree | -Ability to predict a few hydrochar quality parameters<br>-Using an old ML approach in modeling | [24] |
| Sludge, rice straw, distiller's grains, rice straw, straw, microalgae | Support vector machine, artificial neural network, random forest | 226 | CHO analysis, total N content in biomass, ash content, heating temperature, heating rate, residence time, reaction pressure | Process yield, hydrochar yield, hydrochar ash content, hydrochar pH, total N content in hydrochar, total P content in hydrochar | -Ignoring some influential input features<br>-Ability to predict a few hydrochar quality parameters | [25] |

ML technology has been used in previous studies to predict the properties of hydrochar produced from various biomass feedstocks and processing parameters. However, these studies have been limited to non-lignocellulosic biomass and have covered only a few indicators of hydrochar quality. To address these limitations, the present study focuses on developing an inclusive ML model to predict the quantity and quality of hydrochar based on diverse lignocellulosic biomass types and various reaction processing parameters. The novelty of this work lies in the comprehensive experimental database gathered, which covers a wide range of biomass types and processing parameters, and the use of advanced ML techniques to predict the properties of hydrochar with high accuracy. In particular, this study employs decision tree regression (DTR) and support vector regression (SVR) models to predict hydrochar quality and quantity and an evolutionary algorithm to optimize the input variables based on cost functions provided by the selected model. Additionally, a feature importance analysis is conducted to identify the most critical input features that affect the properties of hydrochar. To the best of our knowledge, this study is the first to apply such an inclusive ML approach to predict the properties of hydrochar produced from lignocellulosic biomass feedstocks under various reaction processing parameters.

The results of this study have significant implications for promoting economically viable and environmentally sustainable waste-to-energy systems in urban areas. By developing an effective model to characterize hydrochar produced from various biomass sources under varying reaction processing parameters, the current study can contribute to implementing large-scale HTC systems in industrial settings. Using machine learning to predict the quality and quantity of hydrochar can help optimize the energy and material production potential of biomass wastes while minimizing their negative environmental impacts. This research provides a valuable framework for achieving more efficient and sustainable waste-to-energy systems that benefit the economy and the environment.

## 2. Study Procedures

### 2.1. Research method

The research method used in this study is shown in Figure 1. The first step was to search published research articles on biomass HTC and download the most relevant ones. In the second step, the required data for modeling the process were selected, and the eligible research studies were carefully evaluated to extract the required information. Data mining methods and mechanistic arguments were used to analyze the collected data. Two ML models (SVR and DTR) were developed using the collected dataset to simulate biomass HTC. Following the determination of the best model, a feature importance analysis (SHAP analysis) was conducted to analyze how each independent input feature impacts the corresponding output parameter.

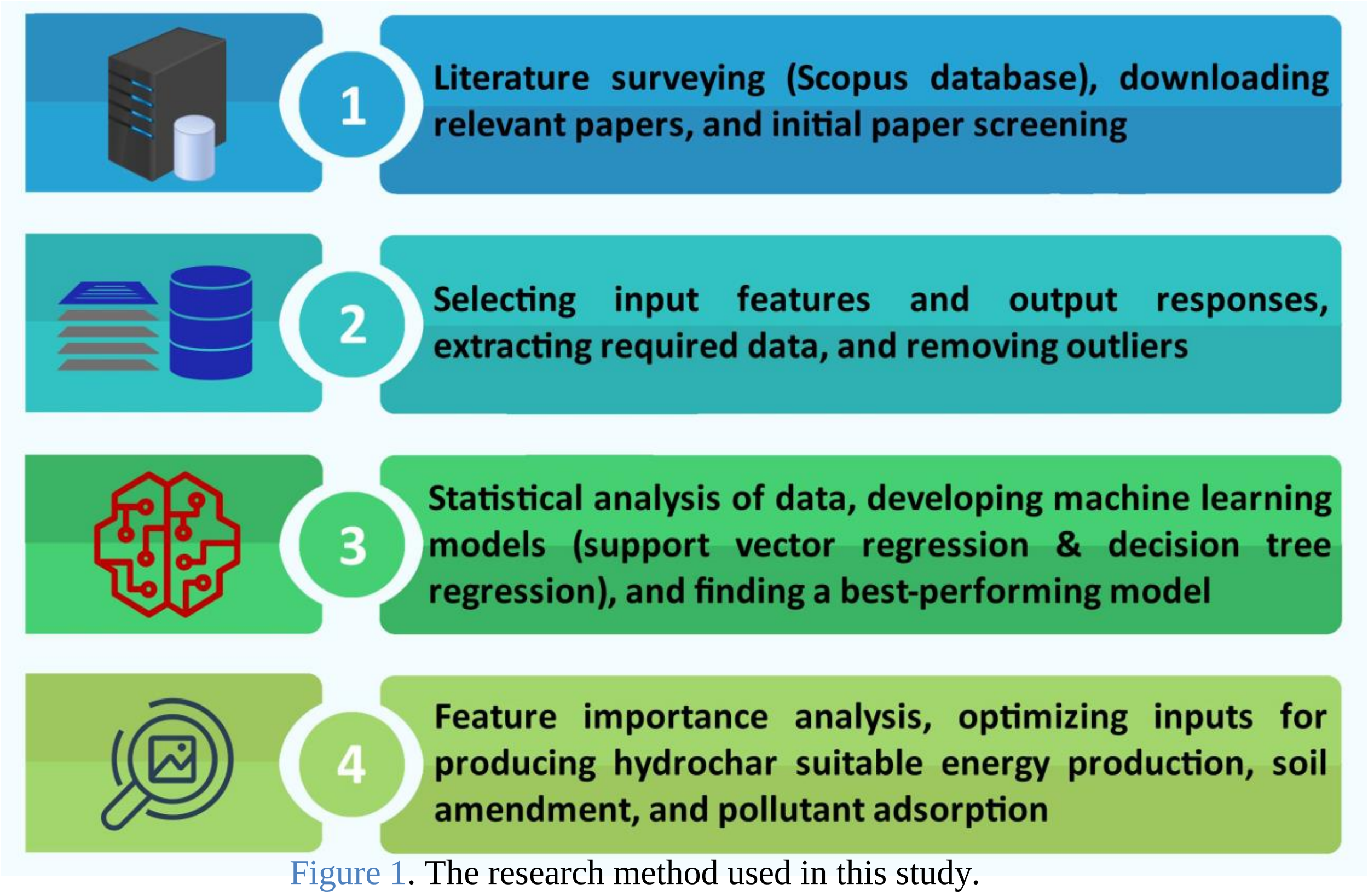

Figure 1. The research method used in this study.

### 2.2. Data collection

The relevant publications on biomass HTC were found by searching the Scopus database. The database was searched for the keywords "biomass" and "HTC" or "hydrothermal carbonization" in the title, abstract, and keywords of the articles. There were 1846 published papers related to biomass HTC in the database. A total of 256 relevant papers were downloaded for further processing. As a next step, the downloaded articles were evaluated and screened to determine whether they contained the data needed. Data patterns were manually extracted using the tables or graphs of the selected papers. The data patterns reported in the graphs were extracted using plot digitizer software. In this study, biomass characteristics and reaction processing parameters are independent input parameters. There are two types of biomass characteristics: proximate and ultimate analyses. The proximate analysis includes the volatile matter, fixed carbon, and ash content (all in wt%) of biomass feedstocks. The ultimate analysis contains carbon, hydrogen, nitrogen, sulfur, and oxygen content (all in wt%) of biomass feedstocks. Temperature (°C), reaction time (min), and water content (wt%) are reaction processing parameters. Hydrochar yield (wt%), hydrochar higher heating value (HHV, MJ/kg), hydrochar proximate analysis (ash content, volatile matter, and fixed carbon), and hydrochar ultimate analysis (C, H, N, S, and O content) are the dependent output parameters. In total, 536 data rows were extracted from 72 high-quality articles. The patterns of collected data can be found in the Supplementary Excel File.

**2.3. Modeling hydrothermal carbonization process**

SVR and DTR were used to develop prediction models for biomass HTC. SVR is an extension of support vector machines that can learn regression functions. This ML model has good generalization capabilities and great potential to handle large feature spaces [26]. The SVR algorithm searches for the global solution using convex quadratic programming optimization theory while avoiding trapping in a local optimum [27]. The SVR algorithm transforms a lower-dimensional nonlinear database (originally n-dimension) into a higher-dimensional space. The goal is achieved by defining an optimum margin to separate the data using a hyperline or a hyperplane. The separating hyperplane or hyperline can be defined using a margin-based loss function (kernel), such as linear, polynomial, and exponential functions [28]. The nonlinear relationship between predictors and targets can be modeled by fitting the data to an appropriate separating hyperline or hyperplane and minimizing the error distance between data points and the defined separator line or plane [29]. With either a small sample size or a large sample size, SVR can demonstrate reliable performance. Using proper kernel functions, the SVR model avoids "overlearning" when solving nonlinear problems [27].

The DTR technique is a nonparametric approach to mining data or developing predictive models in supervised ML [30]. This ML model comprises three main parts, including a root node (whole dataset), a group of internal nodes (splits), and a group of terminal nodes (leaves). The learning process starts from a node at the top of the tree (root node), and then the tree from the root node splits into binary branches (top-down approach). The newly created nodes (interior nodes) follow the splitting data. This separation represents classes or defines data features or decision rules during the process. The process continues to grow until the splitting data reach its optimum model depth, which is the specific stopping criteria (leaf nodes) [30,31]. To find the best-fitting model, DTR iteratively divides the data into smaller segments using a greedy mechanism [32]. Besides modeling complicated phenomena, this ML model can provide details regarding the interpretability of fitted models.

All input and output features were standardized before introducing them to ML models. Standardization was performed by subtracting each data point from the mean and dividing by the standard deviation. The training used 80% of the whole dataset while testing used 20% of the remaining dataset. During the training phase, the k-fold technique was used to develop models. This resampling technique divides data patterns arbitrarily into k parts. k-1 folds are used for training, while the remaining part is used for testing. The training and testing phases are repeated until all folds have participated. The final predictive error is determined by the mean values of each observation [33]. This study applied five-fold cross-validation. The developed models were evaluated during training and testing using three statistical criteria. The statistical criteria were coefficient of determination ($R^2$), mean absolute error (MAE), and root-mean-square error (RMSE) (Eqs. 1–3). It should be noted that each output was analyzed using an exclusive model.

$$R^2 = 1 - \frac{\sum_{i=1}^{n}(y_p - y_a)^2}{\sum_{i=1}^{n}(y_a - \bar{y})^2} \quad (1)$$

$$RMSE = \sqrt{\frac{1}{n}\sum_{i=1}^{n}\left(y_p - y_a\right)^2} \quad (2)$$

$$MAE = \frac{1}{n}\sum_{i=1}^{n}\left|y_a - y_p\right| \times 100 \quad (3)$$

where $y_p$ and $y_a$ stand for the ML-predicted and true data, respectively, $\bar{y}$ denotes the average of the data, and $n$ shows the number of data points.

## 2.4. Process optimization

The HTC process was optimized by employing an evolutionary algorithm (genetic algorithm) to produce hydrochar suitable for energy production, soil amendment, and pollutant adsorption. The algorithm can determine the optimum input variables and their corresponding responses using the cost functions developed by the selected ML model. This meta-heuristic algorithm is designed to solve combinatorial optimization problems by mimicking the natural biological evolution of species [34]. The genetic algorithm creates a search space using individuals (encoded as chromosomes), where each chromosome represents a possible solution to a particular problem [35]. This optimization method uses reproduction, crossover, and mutation to find reasonable solutions for a given problem [36]. As the algorithm evolves, reproduction creates a new population from old individuals. Two mates are selected from the population based on their fitness values. The fitter strings are likelier to enter the mating pool, and the weaker ones are likelier to die off. This process continues, with each old generation replaced by a new one. Therefore, each population is derived from its predecessor [37].

During the crossover stage of the genetic algorithm, a new offspring is created by selecting genes from the parent chromosomes and recombining their genetic material. This operator allows the algorithm to explore the search space more effectively, increasing the likelihood of new generations possessing better characteristics than the previous ones. The mutation operator is another mechanism that helps the algorithm avoid local optima by exploring new states. By changing a random bit (from 1 to 0 or 0 to 1) in a given string, the mutation operator promotes the development of new chromosomes. Prescribing a small mutation probability value controls the process [38]. The genetic algorithm can use these operations to find a global optimum or near-optimum solution [39]. In this study, the generation population, the probability of mating two individuals, and the probability of mutating each individual at each generation were considered 1000, 0.5, and 0.3, respectively. Table 2 outlines the objectives for optimizing the HTC process for different applications of hydrochar.

**Table 2**. Objectives for optimizing the biomass HTC process for different applications of hydrochar.

| **Item** | **Energy production** | **Soil amendment** | **Pollutant adsorbent** |
|---|---|---|---|
| Hydrochar carbon content (wt%) | Maximize | Not important | Not important |
| Hydrochar hydrogen content (wt%) | Maximize | Not important | Not important |
| Hydrochar nitrogen content (wt%) | Minimize | Maximize | Maximize |
| Hydrochar oxygen content (wt%) | Minimize | Not important | Maximize |
| Hydrochar sulfur content (wt%) | Minimize | Maximize | Maximize |
| Hydrochar volatile matter (wt%) | Minimize | Not important | Not important |
| Hydrochar fixed carbon content (wt%) | Not important | Not important | Not important |
| Hydrochar ash content (wt%) | Minimize | Maximize | Maximize |
| Hydrochar HHV (MJ/kg) | Maximize | Minimize | Minimize |
| Hydrochar yield (wt%) | Maximize | Maximize | Maximize |
| Hydrochar carbon content (wt%) | Maximize | Not important | Not important |

## 2.5. Feature importance analysis

Even though ML techniques can potentially model complicated phenomena, their black-box nature makes interpreting the results challenging [40]. Therefore, the results of any ML model should be explained with advanced tools. Two popular approaches to overcoming the challenge of interpreting ML models are partial dependence plots and SHAP analysis. The SHAP method provides a way to measure the impact of each input feature on the output response of an ML model [41]. By quantifying the order of importance, mean marginal contribution (SHAP amount), and direction of effect (positive or negative) of each independent input parameter on dependent output responses [42], SHAP analysis can be used to interpret the developed model either locally or globally. Beeswarm, bar, and heatmap plots are often used to summarize the significance and contribution of each input feature to a dependent output parameter [43].

The SHAP technique offers several advantages over other methods of interpreting ML models. One key advantage is its ability to capture interactions between features, which can help identify more complex data patterns. Additionally, SHAP values are model-agnostic, meaning they can be applied to any model type, regardless of its underlying architecture. Overall, SHAP analysis is a powerful tool for interpreting ML models and understanding the impact of input features on model predictions. By using SHAP analysis in combination with other interpretability techniques, researchers can gain valuable insights into the underlying mechanisms of their models and improve their understanding of complex phenomena.

It is worth mentioning that the statistical analysis of the collected data was performed using the Origin data analysis software. The modeling, optimization, and feature importance analysis were conducted using Python.

## 3. Results and discussion

### 3.1. Descriptive analysis

The compiled research datasets were analyzed statistically to better understand the variables considered in biomass HTC. Some statistical indicators for all the descriptors and responses are summarized in Table S1 (Supplementary Word File). The dataset contains a variety of ultimate and proximate analyses, illustrating the diversity of biomass feedstocks used in published studies. Based on the collected dataset, the carbon content of biomass feedstocks ranges from 22.65–63.82 wt% dry-basis, while their volatile matter varies from 47.38 to 93.42 wt% dry-basis. Carbon content and volatile matter are high in selected biomass, making them suitable for alternative biofuels with enhanced energy density. There are acceptable distributions for hydrogen and oxygen, ranging from 2.9 to 8.1 wt% (with a median value of 6.09 wt%) and 10.5 to 60.5 wt% (with a median value of 40.64 wt%), respectively.

The selected biomass feedstocks are suitable for hydrochar production as their ash content ranges from 0.16 to 49.85 wt%. Biomass feedstocks with higher ash content can produce hydrochar with a higher content of inorganic compounds, such as calcium (Ca) and magnesium (Mg), which can be used as fertilizers. Notably, hydrochar has been found to enhance the effectiveness of fertilizers by reducing nutrient loss through leaching and runoff, as reported in previous studies [44]. The reaction time and operating temperature are highly variable, ranging from 100 to 375°C and 5 to 600 minutes, respectively. The wide range of data included in the dataset effectively covers diverse reaction processing parameters reported in various publications. This comprehensive dataset enables machine learning models to generate highly generalizable results.

The output responses are the ultimate and proximate analyses, heating value, and yield of the resultant hydrochar. Hydrochar carbon content ranges from 19.11 to 78.2 wt%, whereas volatile matter and ash content range from 16.79 to 93.26 wt% and 0.0 to 72.02 wt%, respectively.

As a result of the wide variation in feedstock characteristics and reaction processing parameters, there is a large variation in the hydrochar properties derived from biomass HTC. The maximum HHV of hydrochar is 36.62 MJ/kg, demonstrating its high calorific value and good energy density. Statistical analysis results indicate that the distribution of input and output variables is generally within reasonable bounds (neither dispersed nor concentrated), making the collected dataset suitable for developing ML models.

Biomass HTC is a promising technology that produces hydrochar, a carbon-rich material that is porous and abundant in carbon. Hydrochar has many potential applications, including energy production, environmental protection, soil amendment, and carbon sequestration [45]. The energy density of hydrochar is higher than its parent biomass due to several factors, such as carbon fixation, volatile matter reduction, and ash removal [46]. Hydrochar has a higher calorific value because of its lower O/C and H/C atomic ratios, resulting from the decomposition of cellulose and hemicellulose, and higher C=C chemical bonds due to the promotion of condensation and aromatization reactions [47,48]. Moreover, the high surface area and porosity of hydrochar make it an excellent adsorbent for removing heavy metals, phosphorus and nitrogen, dyes, and emerging contaminants from aqueous environments [46]. The adsorption of contaminants on hydrochar surfaces occurs through various mechanisms, such as electrostatic attraction, ion exchange, surface complexation, and precipitation. Hydrochar has the potential to be a low-cost alternative for the remediation of contaminated soil and water, and its production contributes to reducing greenhouse gas emissions and mitigating climate change.

### 3.2. Spearman correlation analysis

The nonparametric Spearman correlation was used to identify the relevance of two variables in biomass HTC. In this analysis, rank values from the dataset are applied with a monotonic function to assess the preliminary relationship between the two parameters. Specifically, the ranking is conducted in a way where "+1" goes to the highest number in each column, and "-1" goes to the lowest number in each column. Depending on the surveyed features, the Spearman correlation coefficient ranges from -1 to +1, with -1 indicating a complete negative (low impact) and +1 indicating a complete positive. This ranked coefficient is calculated as follows (Eq. 4):

$$r = 1 - \frac{6\sum_i d_i^2}{n(n^2 - 1)} \quad (4)$$

where $r$ stands for Spearman's rank correlation coefficient, $d_i$ represents the difference between the two ranks of each observation, and $n$ is the quantity of observation. The Spearman correlation matrix relating input/output parameters in biomass HTC is shown in Figure 2.

Increasing the operating temperature causes secondary decomposition and Boudouard reactions to occur more quickly, generating more gaseous products and decreasing hydrochar yield [49]. The higher ionization degree allows water to permeate the porous media with increasing temperature easily. Hence, this change could facilitate hydrolysis, cleavage, and ionic condensation of biomass macromolecules [50]. It should be noted that hemicellulose and cellulose are completely hydrolyzed at temperatures around 180 and 220 °C, respectively [51]. Unlike the operating temperature, the reaction time has a less pronounced effect on the hydrochar yield. Since longer reaction times usually result in a more severe reaction, a lower hydrochar yield is achieved.

Hydrochar yield and hydrochar HHV are reversely correlated with the water content. Compared to other hydrothermal processing methods (*i.e.*, hydrothermal liquefaction and gasification), HTC generally operates at mild temperatures and pressures. Subcritical conditions reduce water's density and dielectric constant while increasing its ionization constant significantly. These changes lead to higher dissociations of water into acidic hydronium ions ($H_3O^+$) and hydroxide ions ($OH^-$), facilitating acid- and base-catalyzed reactions [52]. Increasing biomass loading (*i.e.*, decreasing water content) aids in promoting the polymerization reaction in hydrochar formation. Generally, polymerization occurs earlier at lower water contents, increasing hydrochar yield [53].

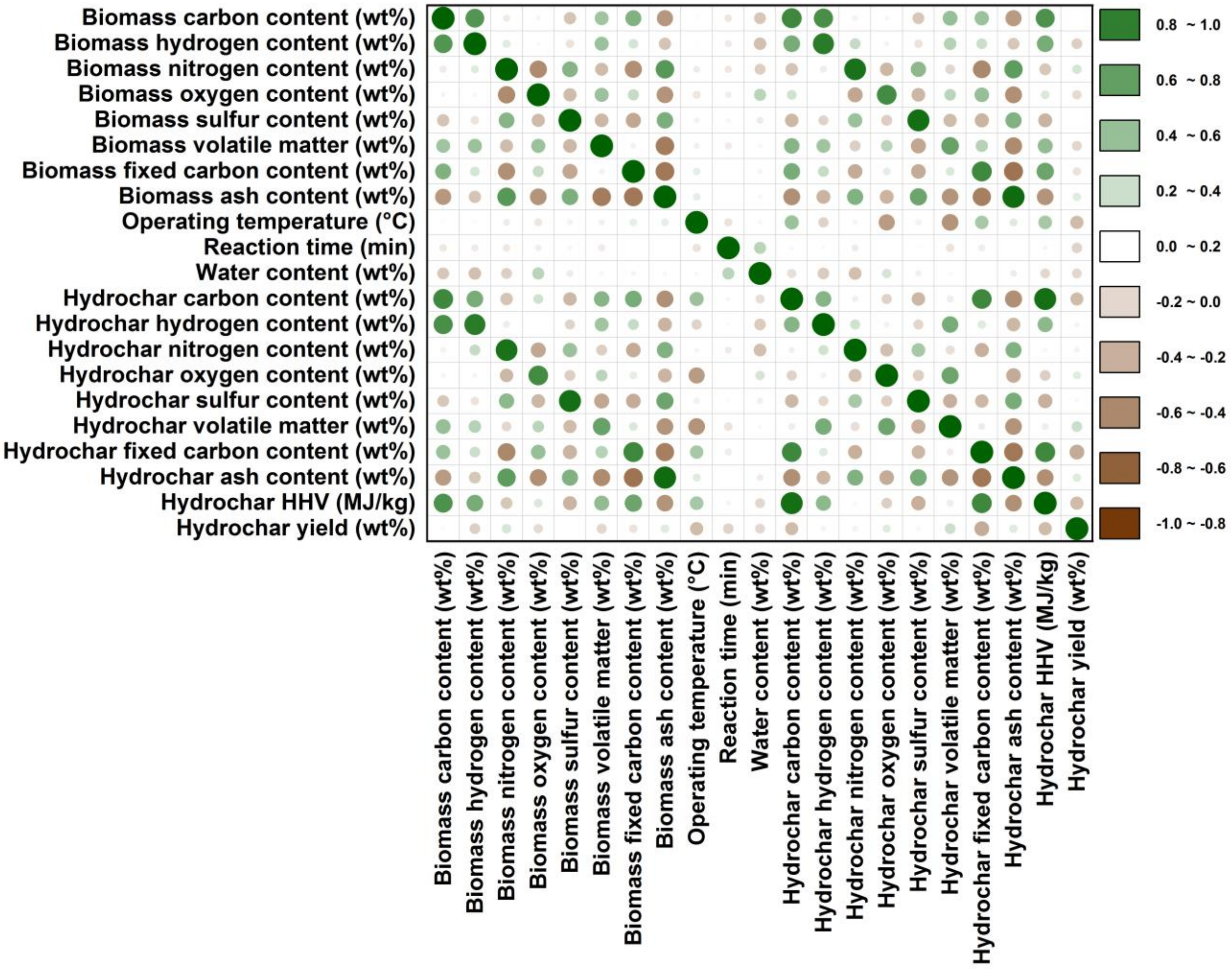

Figure 2. Spearman correlation matrix relating input/output parameters in biomass HTC. Correlations between input descriptors and output targets are shown in the color bar, with green and brown representing positive and negative correlations, respectively.

### 3.3. Factor analysis

Factor analysis is used to understand the information space in a dataset by describing the relationships among the variables. This method identifies the latent variables to explain their positive/negative effects on the output responses. Factor analysis places factors with equal importance in the same direction. Figure 3 depicts the result of factor analysis for modeling data. According to Figure 3A, the first, second, and third factors are responsible for 8.08, 3.55, and 2.71 of the eigenvalues extracted. As shown in Figure 3B, the first three factors account for 68.3% of the total variance. The first three factors could accurately reflect most of the information in the entire dataset. Hydrochar HHV is positively correlated with biomass carbon content since carbon-rich biomass could generate hydrochar with higher carbon contents (Figure 3C). Generally, hydrochar with higher carbon and lower ash contents is preferred for energy generation. Two dominant reactions in HTC enhance the HHV of hydrochar: dehydration and decarboxylation [45].

The presence of inorganic elements in biomass feedstocks with high ash content negatively correlates with the HHV of hydrochar. This result is because energy compounds cannot be formed from inorganic materials such as silicon, potassium, chlorine, and magnesium. Additionally, the ash content of biomass can lead to issues such as fouling, slagging, and corrosion in the combustor [44], while also causing toxic emissions. On the other hand, the HHV of hydrochar is negatively correlated with its water content due to the effect of polymerization reactions, which increase the organic fraction of hydrochar. Indeed, hydrochar has been found to absorb a higher concentration of monomers from the liquid phase due to the acceleration of polymerization reactions. However, it is important to note that these correlations may vary depending on the specific properties of the biomass or the conditions of the reaction process.

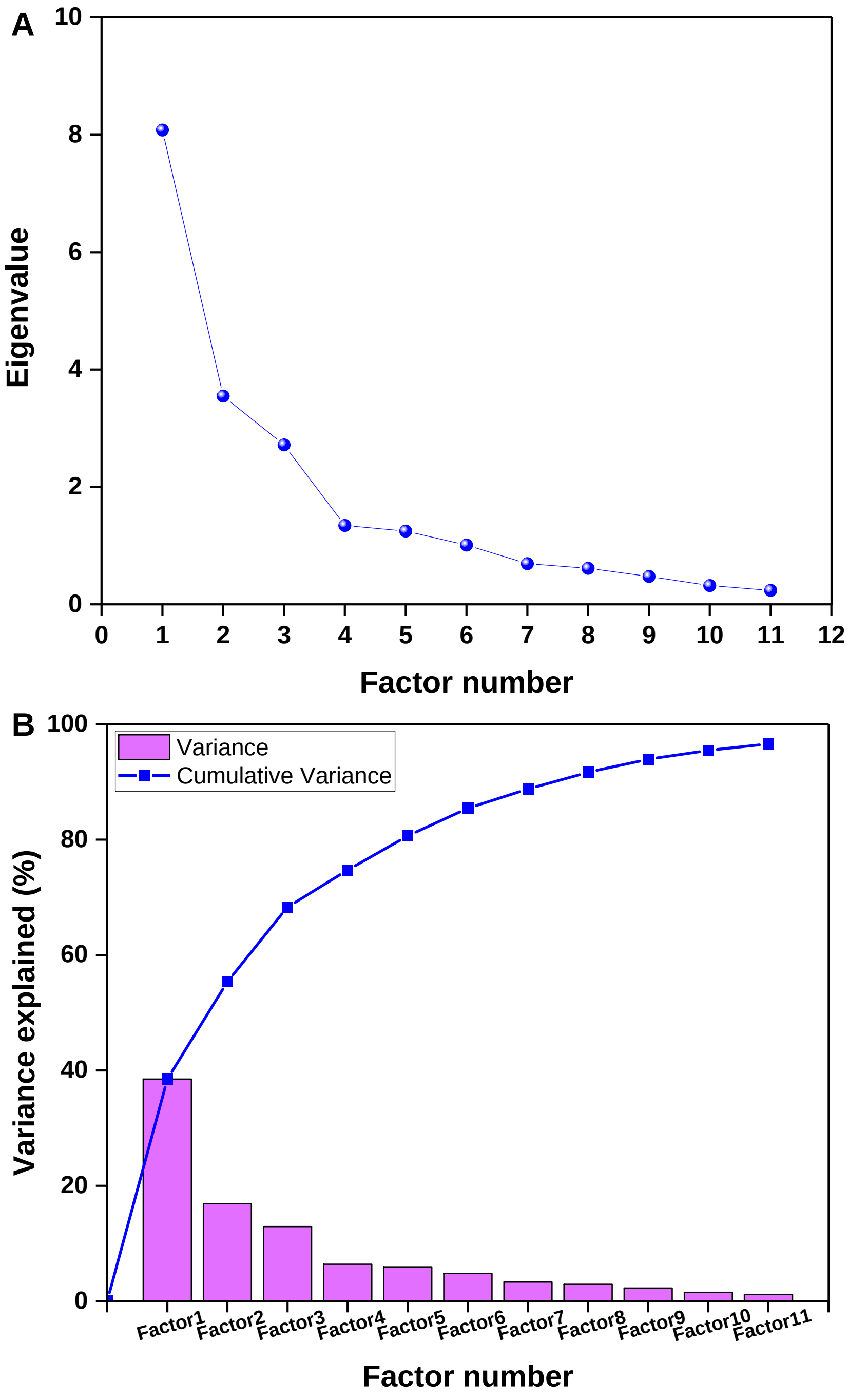

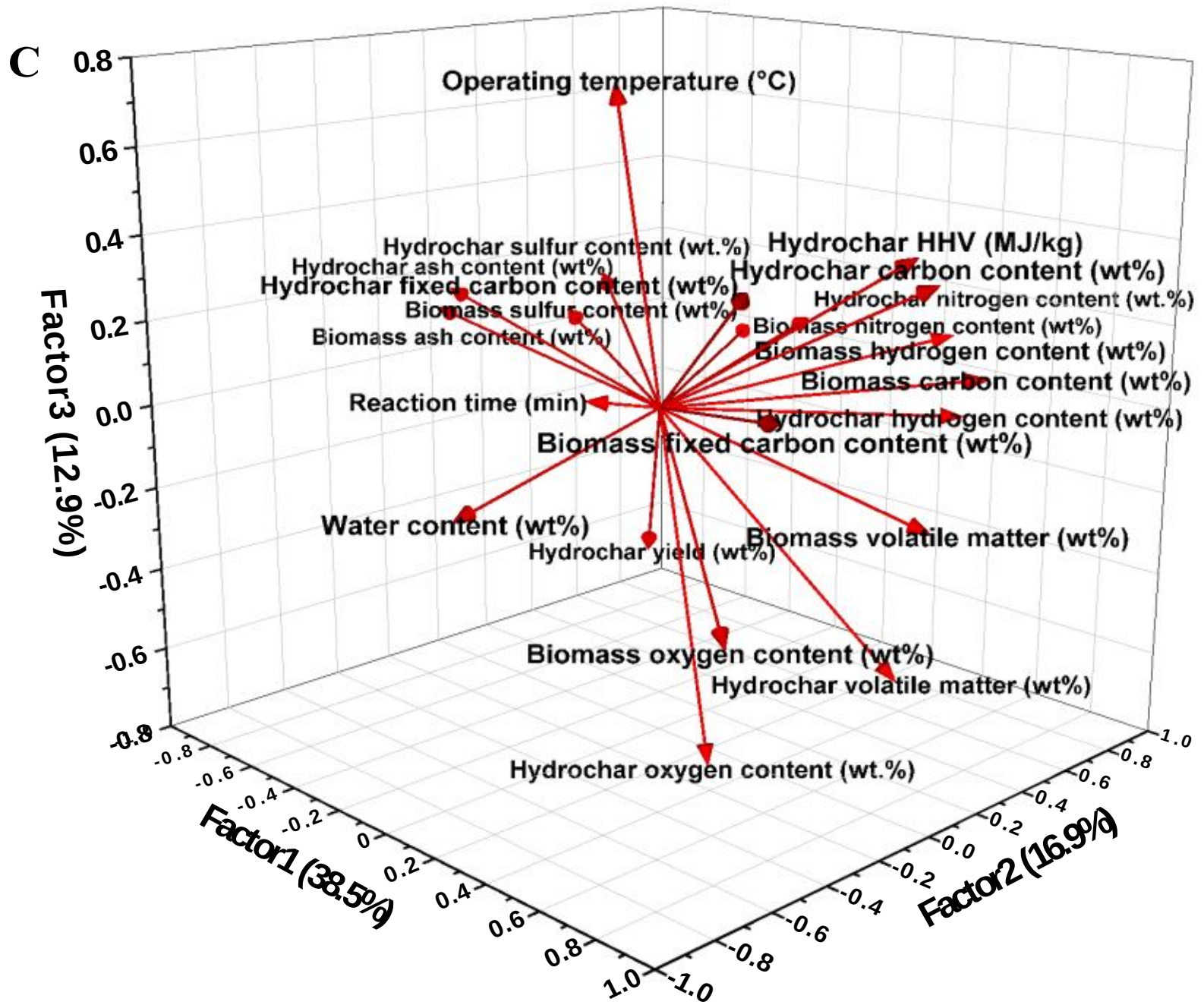

Figure 3. Factor analysis of the dataset used in ML modeling (A) eigenvalue of each extracted factor, (B) variance of each factor, (C) effect of descriptors on responses.

The contour diagram in Figure 4 presents how the most important reaction processing parameters (*i.e.*, temperature, reaction time, and water content) in biomass HTC and hydrochar yield are related. Higher hydrochar yields are achieved when the temperature, reaction time, and water content range between 125 and 175 °C, 25 and 80 min, and 60 and 70%, respectively. The primary function of temperature is to provide enough heat to dissociate chemical bonds with high activity in organic macromolecules. More specifically, the secondary degradation of the solid residues occurs at higher temperatures, expediting the conversion of condensable products into incondensable gaseous products [54]. Thus, higher temperatures decrease hydrochar yield while enhancing syngas yield.

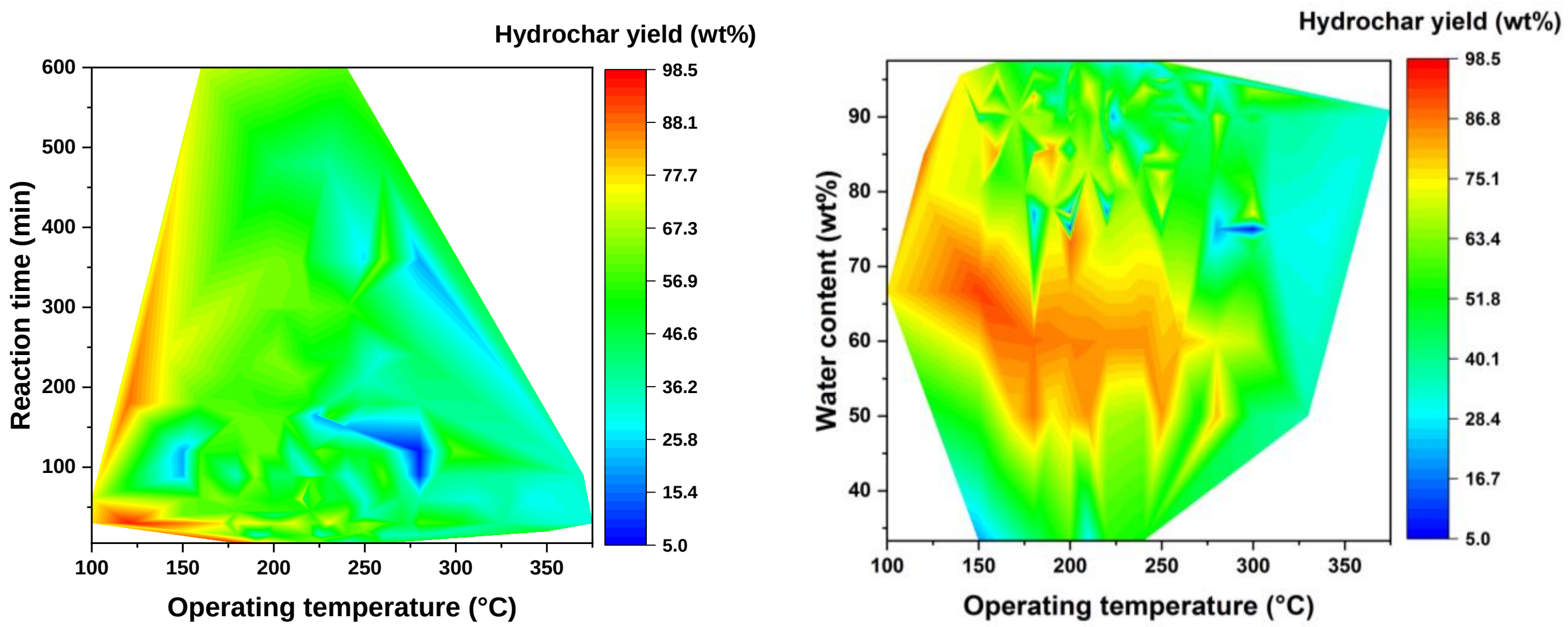

Figure 4. Effect of operating parameters including temperature, reaction time, and water content on hydrochar yield during biomass HTC process.

The graphical representation of the improvement in hydrochar properties during the biomass HTC process is provided by the Van-Krevelen diagram (Figure 5). Biomass and hydrochar have an average H/C ratio of 1.66 and 1.32, respectively. Hydrochar and biomass have an average O/C ratio of 0.42 and 0.66, respectively. Due to the dehydration and decarboxylation reactions occurring during the HTC process, hydrochar has a slightly lower O/C and H/C ratio than the parent biomass. During the mentioned reactions (*i.e.,* dehydration and decarboxylation), the –OH and –COO functional groups are cleaved into $H_2O$ and $CO_2$, respectively. With higher operating temperatures, decarboxylation precedes dehydration, resulting in hydrochar with a smaller O/C atomic ratio than H/C [51].

The demethanation reaction is negligible during the HTC process (Figure 4). Notably, condensation and aromatization reactions can increase the concentration of C=C and C-C chemical bonds in hydrochar, which makes hydrochar a highly energetic densification material. Generally, biomass feedstocks with lower H/C and O/C atomic ratios exhibit a higher degree of condensation and aromaticity during the HTC process. In combustion, these feedstocks lead to less energy loss, smoke emissions, and water vapor formation [51]. Consequently, hydrochar can reduce greenhouse gas emissions when burned or co-burned with coal-like fuels.

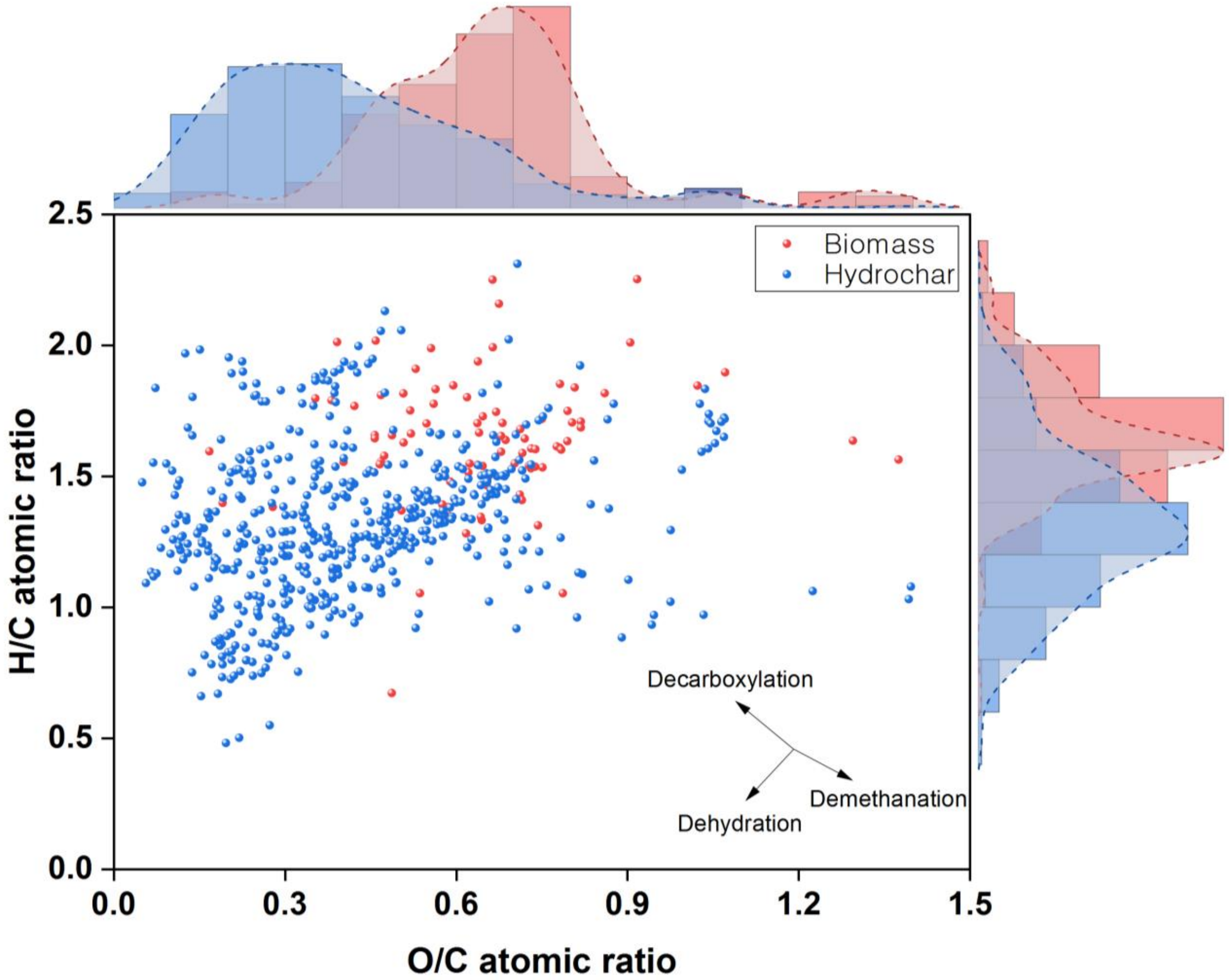

Figure 5. van Krevelen diagram in transforming biomass to hydrochar during the HTC process.

### 3.4. Machine learning modeling evaluation

Table 3 presents the statistical performance parameters of both the DTR and SVR models in characterizing the biomass HTC process. In both the training and testing phases, the DTR model could perform better than the SVR model, with higher $R^2$ values (close to unity) and lower errors (close to zero). The DTR model could yield $R^2$, RMSE, and MAE values ranging from 0.97 to 0.99, 0.124 to 2.554, and 0.01 to 1.351 in the training phase. In the testing phase, $R^2$, RMSE, and MAE for the DTR model are 0.88–0.99, 0.057–6.848, and 0.024–4.718, respectively. According to these results, the DTR model could predict the biomass HTC process more accurately and reliably than the SVR model. The $R^2$ values of the DTR model in the training phase are higher than in the testing phase, so there is no overfitting. The DTR model could provide the highest $R^2$ values in the training and testing phases for the ash content of hydrochar, followed by the sulfur and nitrogen content of hydrochar. The $R^2$ values of the DTR model for the other output parameters, including the yield of hydrochar and its carbon content, HHV, oxygen content, fixed carbon, hydrogen content, and volatile matter, could be found within an acceptable range. The DTR model could provide the lowest RMSE and MAE scores in the training and testing phases for hydrochar sulfur content, followed by the nitrogen and hydrogen content of hydrochar. In addition, the prediction errors of the DTR model for the other output responses are low enough. Table S2 (Supplementary Word File) presents the optimum hyperparameters for the DTR model.

**Table 3**. Analyzing the performance of models based on statistical parameters.

| Statistical parameter | Model type | Hydrochar yield (wt%) | Hydrochar HHV (MJ/kg) | Hydrochar volatile mater (wt%) | Hydrochar fixed carbon (wt%) | Hydrochar ash content (wt%) | Hydrochar carbon content (wt%) | Hydrochar hydrogen content (wt%) | Hydrochar nitrogen content (wt%) | Hydrochar sulfur content (wt%) | Hydrochar oxygen content (wt%) |
|---|---|---|---|---|---|---|---|---|---|---|---|
| *Training phase* | | | | | | | | | | | |
| $R^2$ | SVR | 0.27 | 0.57 | 0.50 | 0.6 | 0.74 | 0.48 | 0.55 | 0.57 | 0.33 | 0.42 |
| | DTR | 0.97 | 0.97 | 0.98 | 0.98 | 0.99 | 0.99 | 0.97 | 0.99 | 0.99 | 0.98 |
| RMSE | SVR | 16.022 | 4.175 | 10.694 | 11.38 | 17.647 | 10.954 | 0.845 | 1.544 | 0.667 | 9.616 |
| | DTR | 2.554 | 0.856 | 1.896 | 1.53 | 0.753 | 1.296 | 0.179 | 0.124 | 0.023 | 1.539 |
| MAE | SVR | 12.718 | 3.134 | 8.373 | 9.936 | 11.255 | 8.048 | 0.579 | 1.021 | 0.278 | 7.528 |
| | DTR | 1.351 | 0.465 | 1.079 | 0.867 | 0.359 | 0.569 | 0.082 | 0.066 | 0.01 | 0.719 |
| *Testing phase* | | | | | | | | | | | |
| $R^2$ | SVR | 0.13 | 0.65 | 0.59 | 0.61 | 0.78 | 0.37 | 0.43 | 0.42 | 0.47 | 0.49 |
| | DTR | 0.88 | 0.94 | 0.93 | 0.93 | 0.99 | 0.95 | 0.95 | 0.98 | 0.98 | 0.95 |
| RMSE | SVR | 18.72 | 3.978 | 11.331 | 12.86 | 20.967 | 13.46 | 0.894 | 1.855 | 0.335 | 10.206 |
| | DTR | 6.848 | 1.410 | 3.833 | 3.806 | 1.847 | 3.361 | 0.248 | 0.323 | 0.057 | 2.655 |
| MAE | SVR | 14.724 | 3.074 | 8.533 | 9.92 | 14.145 | 10.378 | 0.634 | 1.308 | 0.181 | 8.230 |
| | DTR | 4.718 | 0.994 | 2.876 | 2.734 | 1.278 | 2.477 | 0.192 | 0.218 | 0.024 | 2.001 |

The performance of the developed ML models is further evaluated by plotting predicted values against the actual data. Blue and red dots represent the training and testing data points, respectively. Blue and red shadow bands represent the training and testing regression lines with 95% confidence intervals, respectively (Figure 6). The dash lines are 45-degree lines. Figure 6A-B shows linear relationships between the actual and predicted data, with a tight distribution around the 45-degree line in the DTR model. Furthermore, the shadow band of 95% confidence level is narrower in the DTR model. The DTR model is excellent at generalizing the complex biomass HTC process (Figure 6B). The sulfur and ash contents of hydrochar have the tightest data distribution around the dashed 45-degree line compared to other outputs. In addition, there is an acceptable data distribution around the dashed 45-degree line for the other parameters. However, the developed SVR model has many more points deviating from the experimental literature data compared to the DTR model. Overall, the DTR model could excel at predicting the biomass HTC process due to its flexibility in dealing with various response types and the capability to overcome multi-output variables. The result is also attributed to the reliability and explainability of the developed trees and the production of parallel axes to decision surfaces [55,56]. Because biomass HTC is complex and nonlinear, the most widely used SVR model could not accurately predict the dependent output parameters. Table 4 compares the results obtained in this study to those obtained in previous studies. Compared with the models reported in the literature, the DTR model developed herein could predict more information about HTC-derived hydrochar. The capacity of the DTR for predicting the yield of hydrochar and its carbon content, nitrogen content, and HHV could be comparable with the published studies.

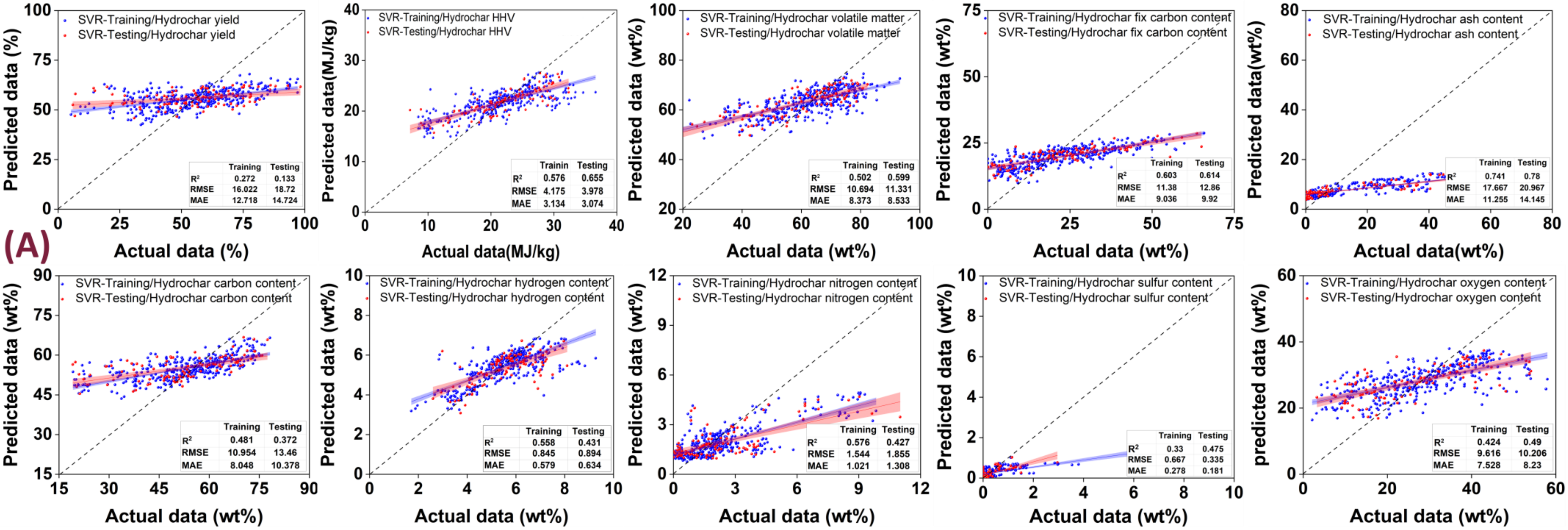

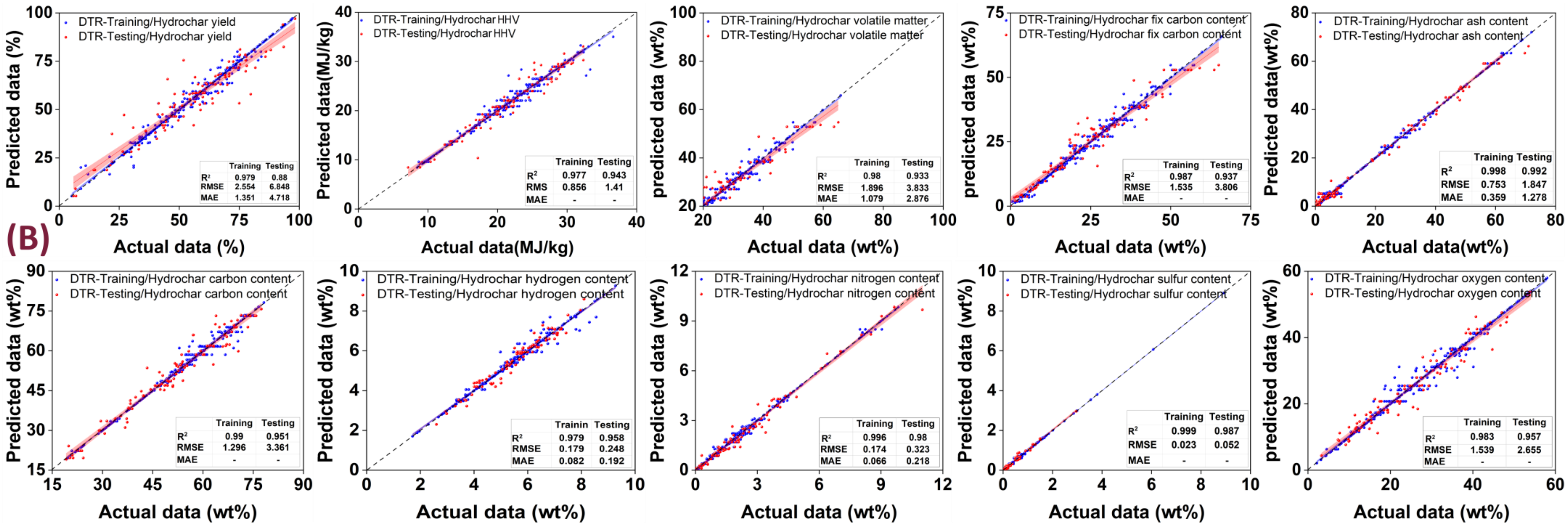

Figure 6. Comparing model-predicted values with actual data. (A) SVR and (B) DTR. Abbreviation: Support vector regression (SVR) and Decision tree regression (DTR)

**Table 4**. Comparing the results obtained in this study to those obtained in previous studies.

| **Ref.** | **Hydrochar yield (%)** | **Hydrochar HHV (MJ/kg)** | **Hydrochar volatile matter (wt%)** | **Hydrochar fixed carbon content (wt%)** | **Hydrochar ash content (wt%)** | **Hydrochar carbon content (wt%)** | **Hydrochar hydrogen content (wt%)** | **Hydrochar nitrogen content (wt%)** | **Hydrochar oxygen content (wt%)** | **Hydrochar sulfur content (wt%)** |
|---|---|---|---|---|---|---|---|---|---|---|
| [18] | RMSE≈7.83 | RMSE≈1.39 | - | - | - | - | - | - | - | - |
| [19] | - | - | - | - | - | - | - | RMSE≈1.39 | - | - |
| [20] | RMSE≈7.05 | RMSE≈7.83 | - | - | - | RMSE≈2.91 | - | - | - | - |
| **This work** | MSE≈6.848 | RMSE≈1.410 | RMSE≈3.833 | RMSE≈3.806 | RMSE≈1.847 | RMSE≈3.361 | RMSE≈0.248 | RMSE≈0.323 | RMSE≈2.655 | RMSE≈0.052 |

### 3.5. Process optimization

The genetic algorithm was applied to determine the optimum biomass composition and reaction processing parameters for producing hydrochar applicable to energy production, soil amendment, and pollutant adsorption (Table 5). For energy production, the optimum hydrochar yield (84.31%) and HHV (34.64 MJ/kg) are obtained when the carbon, hydrogen, nitrogen, oxygen, sulfur, volatile matter, fixed carbon, and ash contents of biomass are 81.23, 7.55, 0.42, 10.78, 0.009, 60.79, 22.02, and 17.18 (all in wt%), respectively. The operating temperature, reaction time, and water content are 227 °C, 26.93 min, and 53.2%, respectively. Under the selected biomass composition and reaction processing parameters, carbon, hydrogen, nitrogen, oxygen, sulfur, volatile matter, and ash content of hydrochar are 69.49, 8.83, 0.296, 3.27, 0.01, 18.81, and 16.7 (all in wt%), respectively.

Regarding soil amendment, the optimum nitrogen, sulfur, and ash content of hydrochar is obtained at 8.51, 8.91, and 68.9 (all in wt%), respectively. These values are observed when the operating temperature, reaction time, and water content are 241 °C, 115.92 min, and 67.95%, respectively. In addition, the carbon, hydrogen, nitrogen, oxygen, sulfur, volatile matter, fixed carbon, and ash content of biomass are 42.26, 4.14, 12.85, 37.09, 3.63, 52.32, 0.12, and 47.55 (all in wt%) respectively. With the chosen biomass composition and reaction processing parameters, hydrochar yield and HHV are 84.91% and 10.22 MJ/kg, respectively.

The optimum carbon, hydrogen, nitrogen, oxygen, sulfur, volatile matter, fixed carbon, and ash content of biomass to reach high-quality hydrochar for pollutant adsorption are 25.06, 7.09, 8.24, 55.96, 3.63, 52.32, 0.12, and 47.55 (all in wt%), respectively. The optimum operating temperature, reaction time, and water content are 299 °C, 19.24 min, and 72.54 wt%, respectively. The optimum nitrogen, oxygen, sulfur, and ash content of hydrochar is 8.95, 57.67, 8.91, and 70.97 (all in wt%), respectively. The hydrochar yield and HHV are 80.40% and 11.23 MJ/kg, respectively, for the chosen biomass composition and reaction processing parameters.

**Table 5**. Optimum biomass composition and reaction processing parameters for producing hydrochar suitable for energy production, soil amendment, and pollutant adsorption.

***Optimum input ranges***

| Hydrochar application | Biomass carbon content (wt%) | Biomass hydrogen content (wt%) | Biomass nitrogen content (wt%) | Biomass oxygen content (wt%) | Biomass sulfur content (wt%) | Biomass volatile matter (wt%) | Biomass fixed carbon content (wt%) | Biomass ash content (wt%) | Operating temperature (°C) | Reaction time (min) | Water content (wt%) |
|---|---|---|---|---|---|---|---|---|---|---|---|
| Energy production | 81.23 | 7.55 | 0.42 | 10.78 | 0.009 | 60.79 | 22.02 | 17.18 | 227 | 26.93 | 53.20 |
| Soil amendment | 42.26 | 4.14 | 12.85 | 37.09 | 3.63 | 52.32 | 0.12 | 47.55 | 241 | 115.92 | 67.95 |
| Pollutant adsorption | 25.06 | 7.09 | 8.24 | 55.96 | 3.63 | 52.32 | 0.12 | 47.55 | 299 | 19.24 | 72.54 |

***Optimum outputs ranges***

| Hydrochar application | Hydrochar yield (%) | Hydrochar HHV (MJ/kg) | Hydrochar carbon content (wt%) | Hydrochar hydrogen content (wt%) | Hydrochar nitrogen content (wt%) | Hydrochar oxygen content (wt%) | Hydrochar sulfur content (wt%) | Hydrochar volatile matter (wt%) | Hydrochar fixed carbon content (wt%) | Hydrochar ash content (wt%) |
|---|---|---|---|---|---|---|---|---|---|---|
| Energy production | 84.31 | 34.64 | 69.49 | 8.83 | 0.296 | 3.27 | 0.01 | 18.81 | - | 16.7 |
| Soil amendment | 84.91 | 10.22 | - | - | 8.51 | - | 8.91 | - | - | 68.9 |
| Pollutant adsorption | 80.40 | 11.23 | - | - | 8.95 | 57.67 | 8.91 | - | - | 70.97 |

### 3.6. Feature importance analysis

ML models are black boxes, so predicting how descriptors affect responses is difficult. Further, it is essential to identify the sensitivity of input-output relationships before developing an ML decision-making framework. The best-performing algorithm (*i.e.*, DTR) is then subjected to SHAP analysis to improve its interpretability by examining the relevance between inputs and outputs. SHAP analysis can provide local interpretability by calculating the significance of input features for each prediction. Additionally, the overall contribution of input descriptors (*i.e.*, global interpretability) is calculated by adding the absolute SHAP values across all datasets. Accordingly, SHAP analysis provides meaningful insights into the output variables from a local and global perspective.

Figure 7 illustrates the SHAP values for each dependent output parameter during the biomass HTC process. It should be noted that input descriptors with high SHAP values are prioritized higher than those with lower values. Input parameters that affect the predicted values positively have a SHAP value greater than zero, while those that affect the predicted values negatively have a SHAP value lower than zero. Each spot in the left-hand plot (*i.e.*, the beeswarm graph), with its color denoting its feature value, represents a single data point in the database. The red and blue points represent high and low original values of the input descriptors, respectively. By increasing the SHAP value, data points turn from blue to red, indicating that the input variable positively influences predicted responses. The middle graph (*i.e.*, the horizontal bar plots) displays the total importance of input descriptors to each target based on the mean absolute SHAP value. The right-hand graph (*i.e.*, heatmap plot) displays a general survey of how the associated response (*i.e.*, f(x) function) modifies as input descriptors vary.

The hydrochar production during the biomass HTC process is primarily affected by biomass ash/carbon content and operating temperature. This result is supported by the mean SHAP values, which indicate that these factors have the highest influence, with values of 7.27, 6.28, and 5.84, respectively. A higher biomass ash content results in a higher positive SHAP value, indicating that it positively influences hydrochar yield. Mineral composition in biomass ash remains in the solid residue product, thereby increasing hydrochar yield [57]. It is worth mentioning that decarboxylation and aromatization reactions in the HTC process are likely to be affected by biomass ash content [20]. In addition, the inorganic fraction of biomass ash could also negatively affect the hydrochar calorific value, hindering the potential application of hydrochar for energy material (Figure 7B). The higher the operating temperature, the more the solid residues decompose, resulting in more incondensable gaseous products. Thus, hydrochar yield is negatively affected by increasing the operating temperature [54].

Figure 7G-J presents the SHAP values of descriptors on the hydrochar characteristics (*i.e.*, ultimate and proximate analyses) during the biomass HTC process. The operating temperature is approximately one of the three most impactful input features, negatively and positively affecting hydrochar properties. For example, the operating temperature positively affects hydrochar fixed carbon content while negatively influencing hydrochar volatile matter. The fixed carbon is correlated to the thermal-stable carbon in hydrochar since it is an original thermally stable component [48]. Increasing the operating temperature generally results in higher carbon and lower oxygen contents in the HTC-derived hydrochar, indicating improved energy characteristics. Neither the hydrochar ultimate nor proximate analyses show obvious monotonic relationships with water content or reaction time.

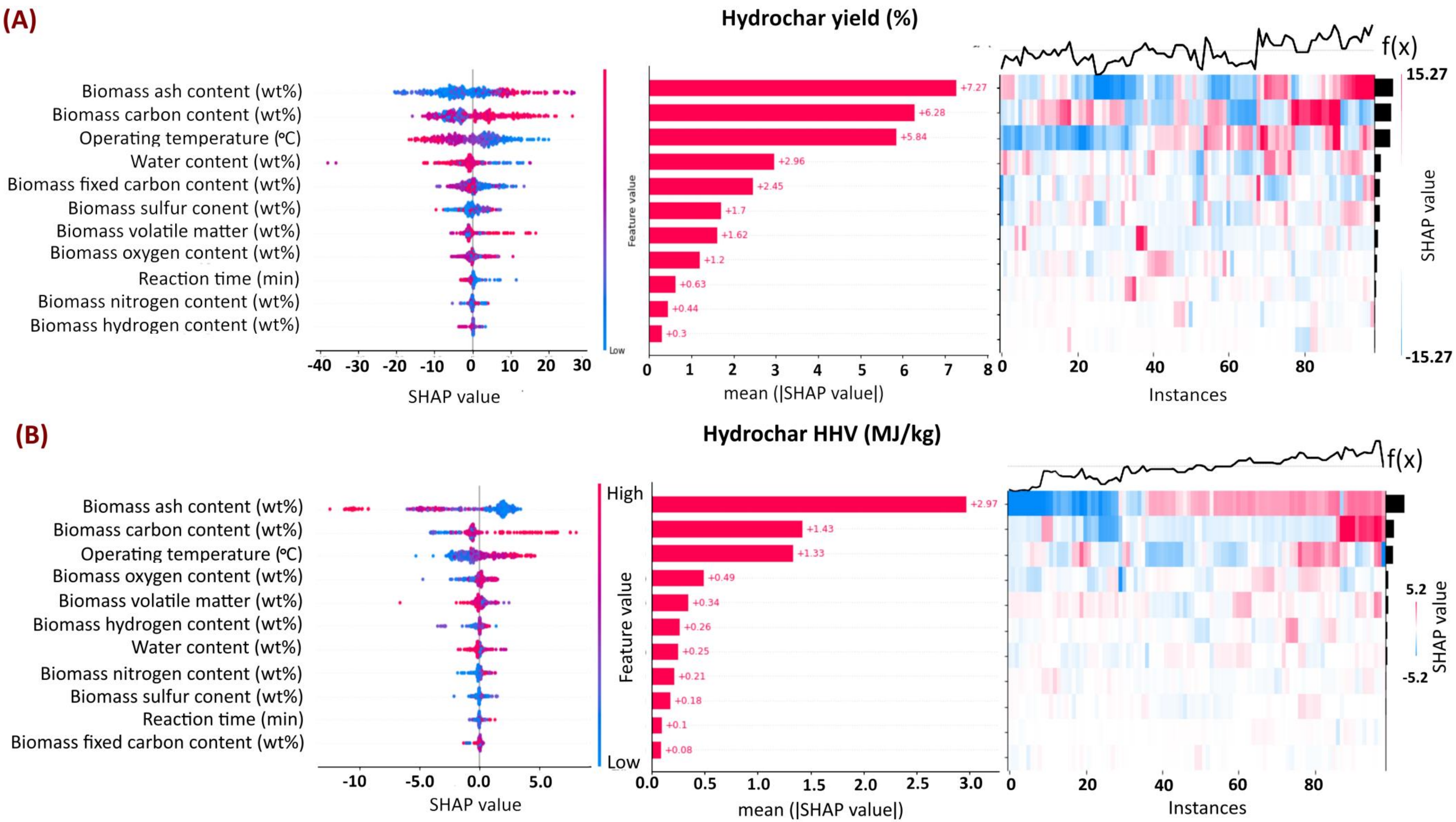

(A)
Hydrochar yield (%)
Biomass ash content (wt%)
Biomass carbon content (wt%)
Operating temperature (ºC)
Water content (wt%)
Biomass fixed carbon content (wt%)
Biomass sulfur conent (wt%)
Biomass volatile matter (wt%)
Biomass oxygen content (wt%)
Reaction time (min)
Biomass nitrogen content (wt%)
Biomass hydrogen content (wt%)
SHAP value
Feature value
Low
mean (|SHAP value|)
+7.27
+6.28
+5.84
+2.96
+2.45
+1.7
+1.62
+1.2
+0.63
+0.44
+0.3
f(x)
Instances
SHAP value
15.27
-15.27
(B)
Hydrochar HHV (MJ/kg)
Biomass ash content (wt%)
Biomass carbon content (wt%)
Operating temperature (ºC)
Biomass oxygen content (wt%)
Biomass volatile matter (wt%)
Biomass hydrogen content (wt%)
Water content (wt%)
Biomass nitrogen content (wt%)
Biomass sulfur conent (wt%)
Reaction time (min)
Biomass fixed carbon content (wt%)
SHAP value
High
Feature value
Low
mean (|SHAP value|)
+2.97
+1.43
+1.33
+0.49
+0.34
+0.26
+0.25
+0.21
+0.18
+0.1
+0.08
f(x)
Instances
5.2
-5.2
SHAP value

(C)

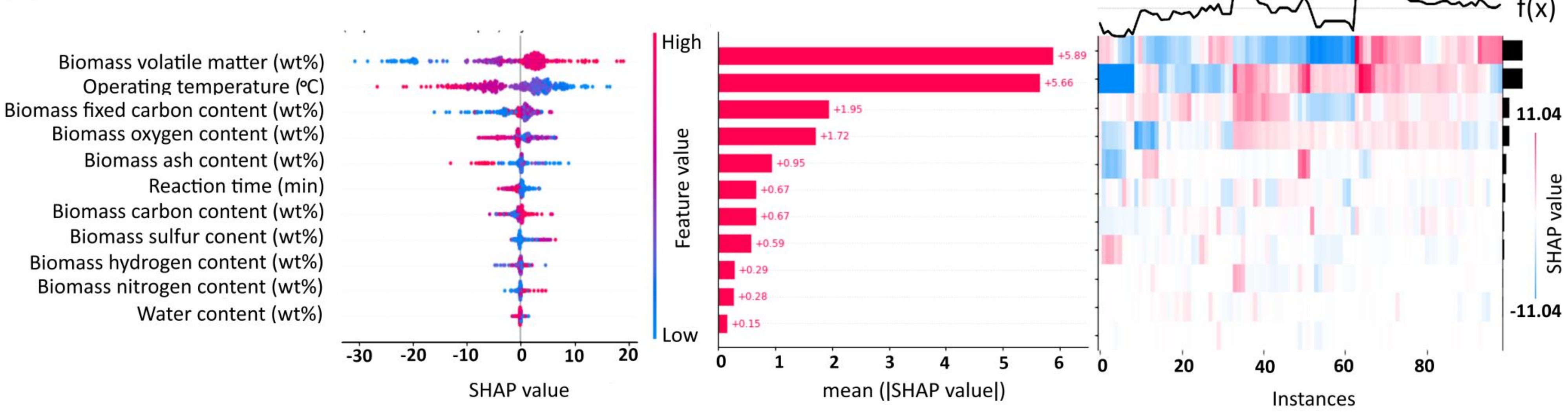
Hydrochar volatile matter (wt%)
Biomass volatile matter (wt%)
Operating temperature (℃)
Biomass fixed carbon content (wt%)
Biomass oxygen content (wt%)
Biomass ash content (wt%)
Reaction time (min)
Biomass carbon content (wt%)
Biomass sulfur conent (wt%)
Biomass hydrogen content (wt%)
Biomass nitrogen content (wt%)
Water content (wt%)
-30 -20 -10 0 10 20
SHAP value
High
Feature value
Low
+5.89
+5.66
+1.95
+1.72
+0.95
+0.67
+0.67
+0.59
+0.29
+0.28
+0.15
0 1 2 3 4 5 6
mean (|SHAP value|)
f(x)
11.04
SHAP value
-11.04
0 20 40 60 80
Instances

(D)

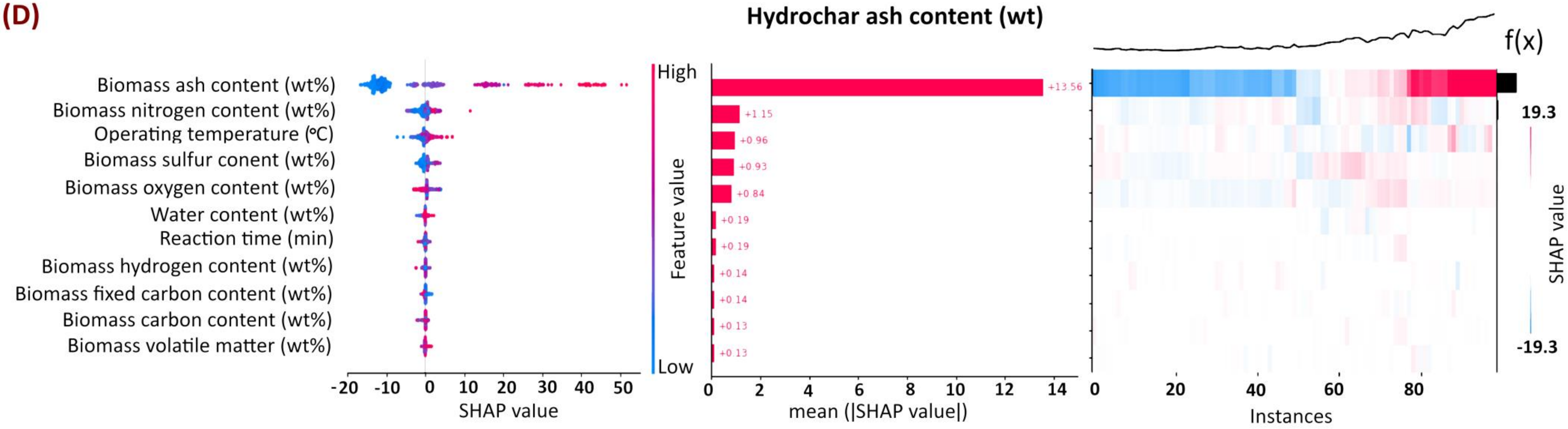
Hydrochar ash content (wt)
Biomass ash content (wt%)
Biomass nitrogen content (wt%)
Operating temperature (℃)
Biomass sulfur conent (wt%)
Biomass oxygen content (wt%)
Water content (wt%)
Reaction time (min)
Biomass hydrogen content (wt%)
Biomass fixed carbon content (wt%)
Biomass carbon content (wt%)
Biomass volatile matter (wt%)
-20 -10 0 10 20 30 40 50
SHAP value
High
Feature value
Low
+13.56
+1.15
+0.96
+0.93
+0.84
+0.19
+0.19
+0.14
+0.14
+0.13
+0.13
0 2 4 6 8 10 12 14
mean (|SHAP value|)
f(x)
19.3
SHAP value
-19.3
0 20 40 60 80
Instances

(E)

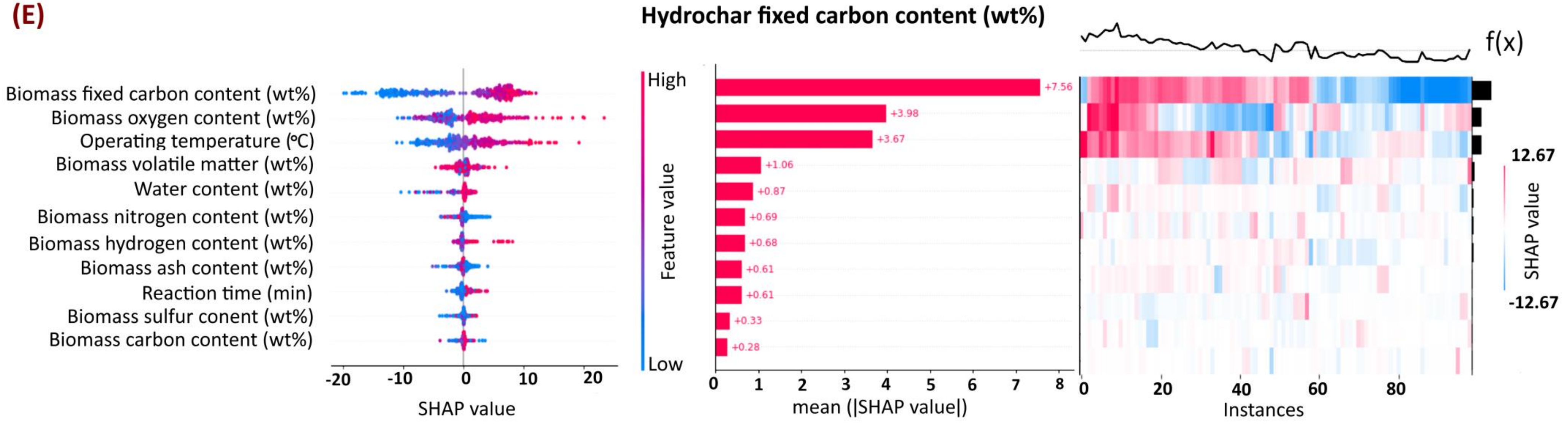
Hydrochar fixed carbon content (wt%)
Biomass fixed carbon content (wt%)
Biomass oxygen content (wt%)
Operating temperature (ºC)
Biomass volatile matter (wt%)
Water content (wt%)
Biomass nitrogen content (wt%)
Biomass hydrogen content (wt%)
Biomass ash content (wt%)
Reaction time (min)
Biomass sulfur conent (wt%)
Biomass carbon content (wt%)
-20
-10
0
10
20
SHAP value
High
Feature value
Low
+7.56
+3.98
+3.67
+1.06
+0.87
+0.69
+0.68
+0.61
+0.61
+0.33
+0.28
0 1 2 3 4 5 6 7 8
mean (|SHAP value|)
f(x)
12.67
SHAP value
-12.67
0 20 40 60 80
Instances

(F)

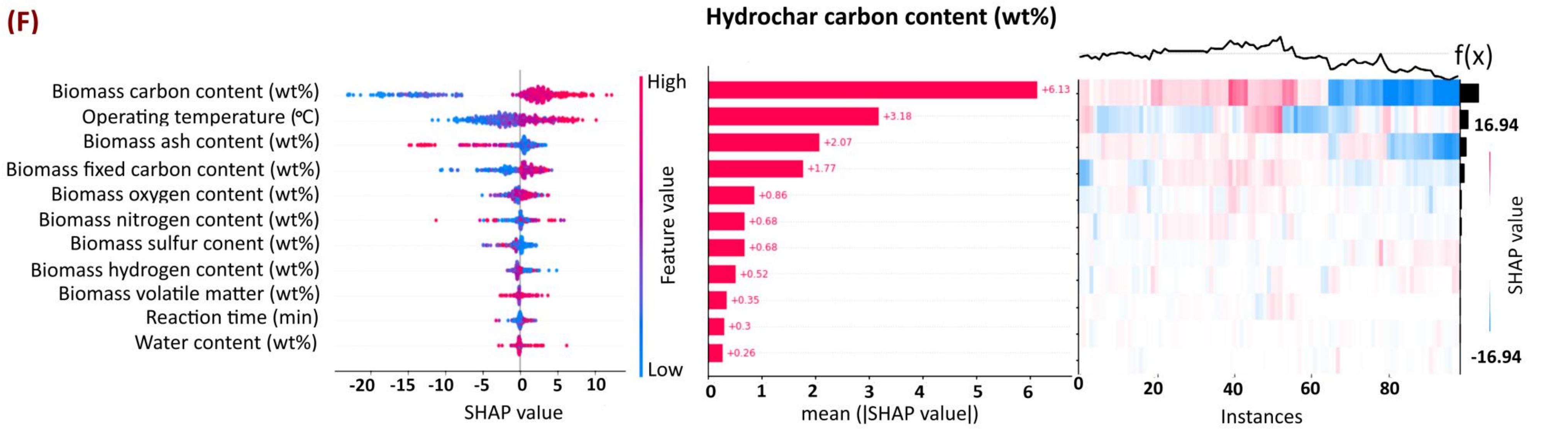
Hydrochar carbon content (wt%)
Biomass carbon content (wt%)
Operating temperature (ºC)
Biomass ash content (wt%)
Biomass fixed carbon content (wt%)
Biomass oxygen content (wt%)
Biomass nitrogen content (wt%)
Biomass sulfur conent (wt%)
Biomass hydrogen content (wt%)
Biomass volatile matter (wt%)
Reaction time (min)
Water content (wt%)
-20 -15 -10 -5 0 5 10
SHAP value
High
Feature value
Low
+6.13
+3.18
+2.07
+1.77
+0.86
+0.68
+0.68
+0.52
+0.35
+0.3
+0.26
0 1 2 3 4 5 6
mean (|SHAP value|)
f(x)
16.94
SHAP value
-16.94
0 20 40 60 80
Instances

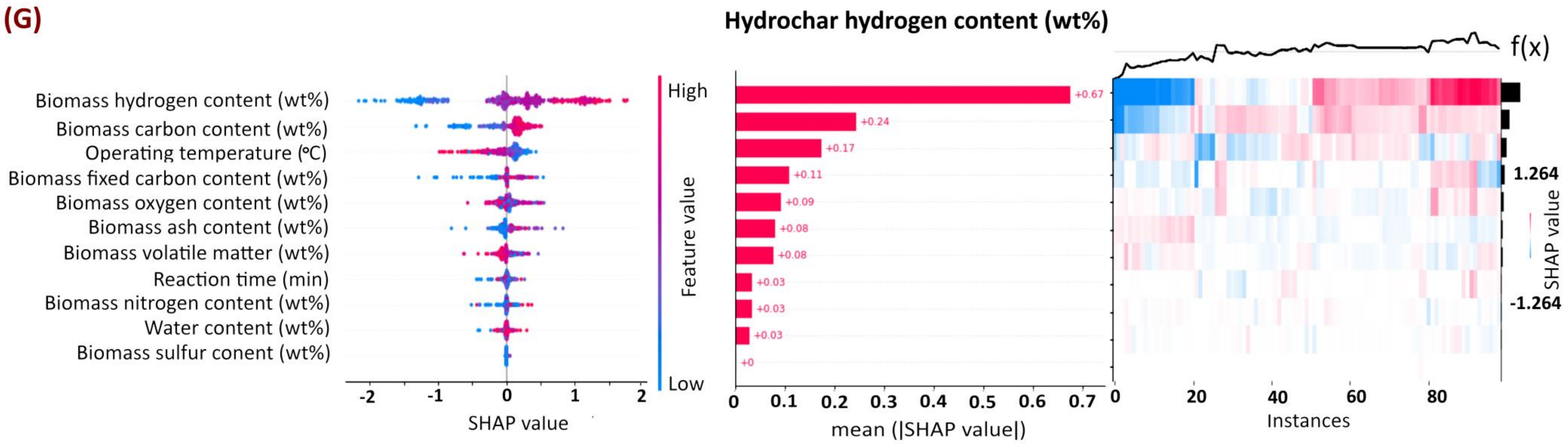
(G)
Hydrochar hydrogen content (wt%)
Biomass hydrogen content (wt%)
Biomass carbon content (wt%)
Operating temperature (°C)
Biomass fixed carbon content (wt%)
Biomass oxygen content (wt%)
Biomass ash content (wt%)
Biomass volatile matter (wt%)
Reaction time (min)
Biomass nitrogen content (wt%)
Water content (wt%)
Biomass sulfur conent (wt%)
SHAP value
High
Feature value
Low
+0.67
+0.24
+0.17
+0.11
+0.09
+0.08
+0.08
+0.03
+0.03
+0.03
+0
mean (|SHAP value|)
f(x)
1.264
-1.264
SHAP value
Instances

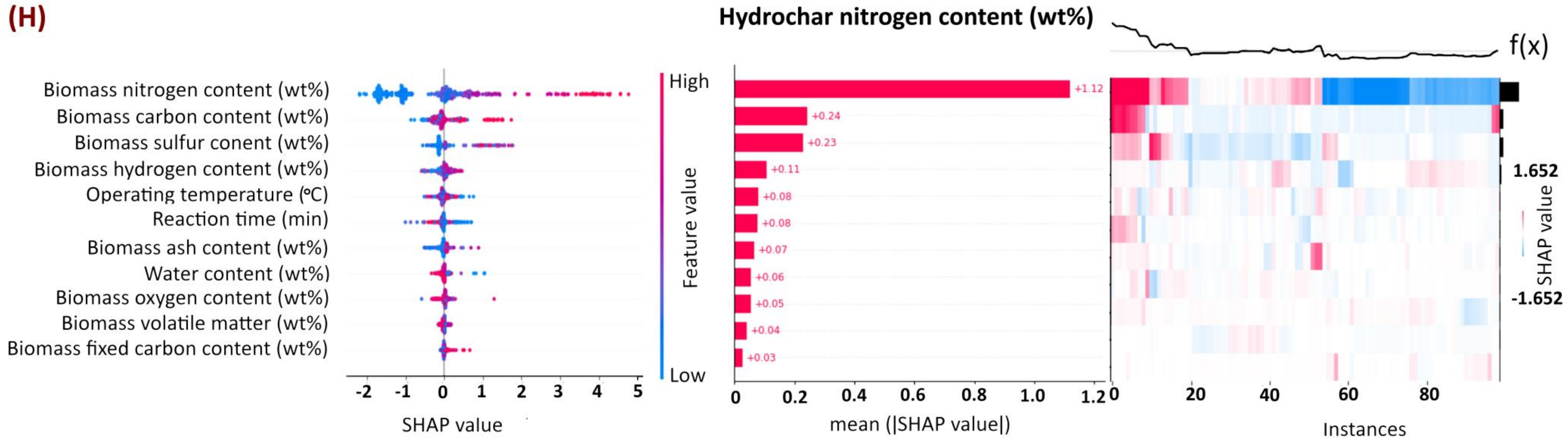
(H)
Hydrochar nitrogen content (wt%)
Biomass nitrogen content (wt%)
Biomass carbon content (wt%)
Biomass sulfur conent (wt%)
Biomass hydrogen content (wt%)
Operating temperature (°C)
Reaction time (min)
Biomass ash content (wt%)
Water content (wt%)
Biomass oxygen content (wt%)
Biomass volatile matter (wt%)
Biomass fixed carbon content (wt%)
SHAP value
High
Feature value
Low
+1.12
+0.24
+0.23
+0.11
+0.08
+0.08
+0.07
+0.06
+0.05
+0.04
+0.03
mean (|SHAP value|)
f(x)
1.652
-1.652
SHAP value
Instances

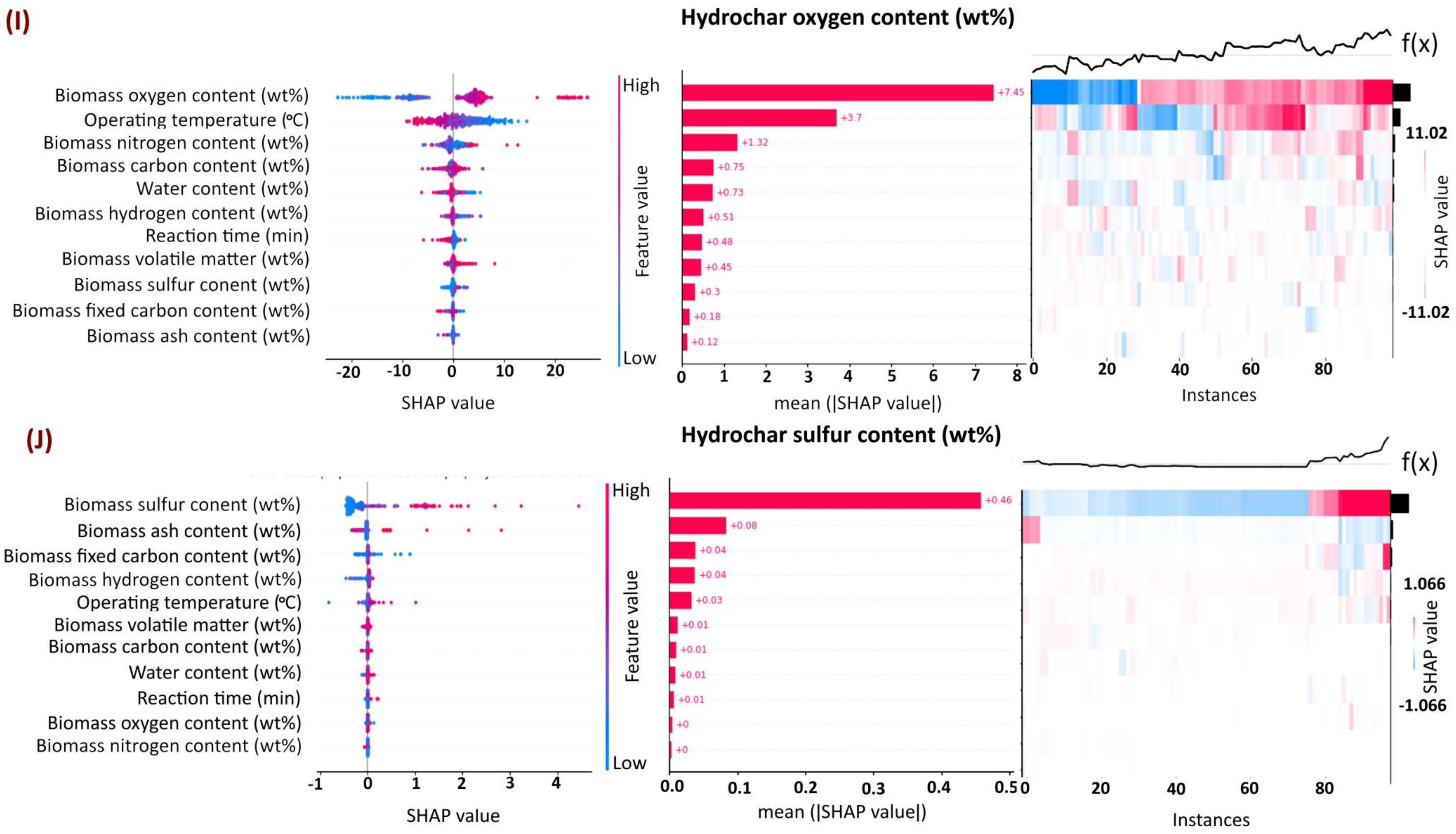

Figure 7. Feature importance analysis (SHAP) of the developed DTR model for predicting hydrochar production during the biomass HTC process.

### 3.7. Limitations and opportunities for future work

While the results obtained in this study are promising, it is still possible to characterize the biomass HTC process with more accurate and practical models. The collected dataset does not include details about biomass ash. Biomass ash constituents can catalyze biomass thermochemical decomposition reactions. As a result, ML models can be improved by knowing more about the ash content of the biomass. In addition, biomass ash compounds mainly remain in the resultant hydrochar, affecting its subsequent use. For example, ash compounds such as inorganics (K, Ca, Na, and Mg) and heavy metals are important when hydrochar is applied to soil amendment. Accordingly, ML technology can help find the best application for hydrochar by providing more detailed information about its ash content.

In addition to water, other solvents (*e.g.*, alcohols) can be employed in the biomass HTC process. Future studies need to develop inclusive models that can cover a variety of solvents used in biomass HTC. Hydrochar with particular properties can also be obtained from the biomass HTC process in the presence of a catalyst. Future research should develop models applicable to both catalytic and non-catalytic biomass HTC processes. To find the right real-world application for hydrochar, it is crucial to model its structural and morphological properties (*i.e.*, crystallinity index, average crystal size, surface morphology, surface area, pore volume, and surface functional groups) using ML technology. High-density hydrochar can be produced by co-processing biomass feedstocks and waste materials (*e.g.*, petroleum waste fuels or polymer wastes). This process can reduce the environmental impact of waste and improve the utilization of biomass because of the synergistic interactions between them. Consequently, future research should develop a more detailed model for co-processing biomass feedstocks and waste materials during the HTC process.

Although biomass HTC has been extensively studied, liquid and gaseous products from this process have received limited attention. To better understand hydrochar formation in biomass HTC, these by-products containing intermediate products must be analyzed more thoroughly. Developing models to predict liquid and gaseous products from biomass HTC is necessary. Water recirculation can significantly reduce water consumption, wastewater generation, and energy recovery efficiency in real-world HTC systems. Controlling pollutant quantity is particularly important in the process of process water circulation. Therefore, a detailed model should be developed to predict the hydrochar properties following various recirculation cycles.

## 4. Conclusions

This study performed an ML model to characterize HTC-derived hydrochar based on the dataset collected from the experimental literature. The collected dataset was statistically analyzed before being used in ML models to understand how the input descriptors affect hydrochar and its physicochemical properties. Two ML algorithms (SVR and DTR) were used to model the quantity and quality of hydrochar (yield, HHV, ultimate analysis, and proximate analysis). The DTR model was more accurate than the SVR model, with an $R^2$ value over 0.88, an RMSE below 6.848, and an MAE below 4.718. The cost functions provided by the DTR model were successfully used in the evolutionary genetic algorithm to determine the optimum input variables and their corresponding responses. The optimization was carried out to produce hydrochar suitable for producing energy, amending soil, and removing pollution.

By analyzing feature importance (*i.e.,* SHAP), key descriptors contributing more to the model's accuracy were identified. Biomass ash content and operating temperature were the two most significant parameters affecting hydrochar production. The ML model developed for biomass

HTC could save time and cost during the experimental phase. The model could provide researchers with an effective tool for finding optimum reaction processing parameters within a limited budget and timeframe. Moreover, the presented model could also aid in analyzing large-scale biomass HTC systems from economic, technological, and environmental perspectives. Monitoring, controlling, and optimizing biomass HTC reactors in real-time could be possible using the best-performing model to enhance hydrochar production yield and quality.

**Acknowledgments**

The authors would like to thank Universiti Malaysia Terengganu under International Partnership Research Grant (UMT/CRIM/2-2/2/23 (23), Vot 55302) for supporting this joint project with Henan Agricultural University under a Research Collaboration Agreement (RCA). This work is also supported by the Ministry of Higher Education, Malaysia, under the Higher Institution Centre of Excellence (HICoE), Institute of Tropical Aquaculture and Fisheries (AKUATROP) program (Vot. No. 56052, UMT/CRIM/2-2/5 Jilid 2 (11)). The manuscript is also supported by the Program for Innovative Research Team (in Science and Technology) at the University of Henan Province (No. 21IRTSTHN020) and the Central Plain Scholar Funding Project of Henan Province (No. 212101510005). The authors would also like to extend their sincere appreciation to the University of Tehran and the Biofuel Research Team (BRTeam) for their support throughout this project.